\documentclass{ws-ijns}
\usepackage{graphicx}
\usepackage[english]{babel}
\usepackage[super,sort,compress]{cite}
\usepackage{color}
\begin{document}

\catchline{}{}{2022}{}{}

\markboth{Avola et al.}{Human Silhouette and Skeleton Video Synthesis through Wi-Fi signals}

\title{Human Silhouette and Skeleton Video Synthesis through Wi-Fi signals}

\author{Danilo Avola\footnote{Corresponding author.}, Marco Cascio, Luigi Cinque, and Alessio Fagioli}
\address{Department of Computer Science, Sapienza University of Rome, \\Via Salaria 113, Rome, 00198, Italy\\ avola,@di.uniroma1.it\\cascio@di.uniroma1.it\\cinque@di.uniroma1.it\\fagioli@di.uniroma1.it}

\author{Gian Luca Foresti}
\address{Department of Computer Science, Mathematics and Physics,\\ University of Udine, Via delle Scienze 206, Udine, 33100, Italy\\gianluca.foresti@uniud.it}

\maketitle

\begin{abstract}
The increasing availability of wireless access points (APs) is leading towards human sensing applications based on Wi-Fi signals as support or alternative tools to the widespread visual sensors, where the signals enable to address well-known vision-related problems such as illumination changes or occlusions. Indeed, using image synthesis techniques to translate radio frequencies to the visible spectrum can become essential to obtain otherwise unavailable visual data. This domain-to-domain translation is feasible because both objects and people affect electromagnetic waves, causing radio and optical frequencies variations. In literature, models capable of inferring radio-to-visual features mappings have gained momentum in the last few years since frequency changes can be observed in the radio domain through the channel state information (CSI) of Wi-Fi APs, enabling signal-based feature extraction, e.g., amplitude. On this account, this paper presents a novel two-branch generative neural network that effectively maps radio data into visual features, following a teacher-student design that exploits a cross-modality supervision strategy. The latter conditions signal-based features in the visual domain to completely replace visual data. Once trained, the proposed method synthesizes human silhouette and skeleton videos using exclusively Wi-Fi signals. The approach is evaluated on publicly available data, where it obtains remarkable results for both silhouette and skeleton videos generation, demonstrating the effectiveness of the proposed cross-modality supervision strategy. 
\end{abstract}

\keywords{Human silhouette; video synthesis; Wi-Fi signal; skeleton.}

\begin{multicols}{2}

\section{Introduction}
In recent years, deep neural networks have been crucial to successfully model and solve diverse real-world applications spanning over, for example, object detection \cite{pankaj2020neural,nogay2020detection} and recognition\cite{acharya2018deep,ahmad2019novel}, health-care \cite{acharya2018automated,nogay2020machine}, forecasting \cite{torres2018scalable,pedro2021experimental}, kinematics \cite{Schwan2021ATM,jesus2021modified}, information filtering \cite{wei2017collaborative,guilherme2020deep}, and image synthesis \cite{zhu2017unpaired,fang2020identity} based tasks. Amid these relevant tasks, one of particular interest is image synthesis, which refers to the creation of a new image using different types of sources as an image description. To this end, the state-of-the-art proposed several promising deep learning methods, including, but not limited to, text-to-image \cite{zhang2017stackgan,zhu2020cookgan}, sketch-to-image \cite{sangkloy2017scribbler,chen2018sketchy}, and image-to-image translation \cite{isola2017image,mejjati2018unsupervised}. Among the existing deep neural networks, the generative adversarial network (GAN) \cite{goodfellow2014generative} proved to generate reasonable, high-resolution, and realistic synthetic images \cite{tsinghua2017survey,mejjati2018unsupervised}. Indeed, its generative power has ensured impressive results in many image synthesis and application-driven tasks, e.g., image dataset augmentation \cite{zhang2020pac-gan}, anomaly detection \cite{shin2020convolutional}, industrial simulation \cite{evgeny2021conditional}, 3D object transformation \cite{ysbrand2019itergans}, pluralistic image completion \cite{Zheng2019pluralistic}, or human-related image generation \cite{song2019unsupervised}. In particular, the latter represents one of the most interesting future research directions. In general, works performing synthesis of human-related images exploit an initial image description that usually consists of text-based or visual information \cite{zhou2019text, wang2020state}. However, since traditional methods based on visual data are not helpful under challenging circumstances in which occlusions, smoke, or darkness can limit visibility, wireless signals were also explored to model and give a visual appearance to otherwise unavailable information carried out by radio waves. \cite{guo2020signal,kato2021csi2image} What is more, other than the access to missing data, human privacy is naturally preserved because sensitive information is not collected through the radio medium. Thus, by translating radio frequency information to the visible spectrum, vision-based systems can be enhanced with powerful and exciting capabilities. This domain translation is feasible because both inanimate objects and people in an environment affect the electromagnetic (EM) spectrum at different frequencies. Hence, measurements of the same entity performed at radio frequency and optic ranges can be correlated \cite{kefayati2020wi2vid}. Despite this, specific mathematical operations do not exist to define this duality. Nevertheless, in literature, models capable of inferring a mapping between radio and optical frequencies were introduced in the past. For instance, the ray-tracing for radio waves propagation modeling is motivated by ray optics \cite{yun2015ray}. Consequently, the analogies between the radio and visible EM spectra have made it reasonable to model the wireless signals to solve traditional vision-based tasks through wireless sensing approaches\cite{malekzadeh2019stupefy,xu2017radio}. Another important aspect is that wireless technology can be cost-effective, thanks to the now ubiquitousness of Wi-Fi signals in public and private places. Generally, vision-based applications are constrained to the frame of low-resolution or costly visual sensors, requiring multiple devices strategically located in different places to cover different points of view, even for small limited areas. In contrast, wireless sensing applications can cover vast areas, potentially exploiting the existing networks built through commodity, commercial, and low-cost Wi-Fi devices. Indeed, even the IEEE 802.11 wireless LAN working group is already focusing on WLAN sensing applications \cite{restuccia2021ieee}, and in the near future Wi-Fi might become a real sensing technology. To this end, channel state information (CSI) based methods are the most popular since this measurement can be easily computed via commodity Wi-Fi devices. Specifically, the CSI describes the signal propagation among two access points (APs) and it is an essential starting point to develop wireless sensing applications. In fact, through CSI it is possible to extract signal features, e.g., amplitude or phase, and catch signal variations caused by people or objects along the propagation path. Actually, such radio-based features are already being employed to develop practical Wi-Fi sensing applications \cite{shi2018accurate,li2021amplitude,wang2015phasefi,rao2020dfphasefl}, including human pose estimation synthesizing skeletal visual data \cite{zhao2018human,zhao2018skeletons}. However, human-related data synthesis works are generally based on non-commodity devices or exploit raw CSI measurements. Although effective, they can be improved by leveraging commodity hardware in conjunction with strategies focusing on improving the CSI measurement quality. The latter is usually obtained by removing noise from the received signal through, for example, outliers removal\cite{dang2019novel}, thus resulting in robust signal-based features that can further improve the addressed task.

Motivated by the existing literature, this paper introduces a novel two-branch generative neural network that maps data acquired in the wireless spectrum to the visual domain, following a teacher-student fashion \cite{avola2019master}. Specifically, sanitized Wi-Fi signals amplitudes extracted from CSI measurements are used to synthesize video frames depicting human silhouettes or skeletons performing different poses. Signal amplitudes were chosen since the literature proves that they can discriminate human activities adequately \cite{wu2019twsee,li2019wimotion,shen2019wirim}, supporting the immediate investigation of such radio features to generate new pose-related visual data. In particular, the proposed method exploits cross-modality supervision \cite{arsha2020speech2action} at multiple levels of the network training pipeline to transfer visual knowledge to Wi-Fi data and learn how to synthesize videos, instead of static images, of human silhouettes from signal-based features. What is more, by leveraging 3D-GAN \cite{wu2016learning}, long short-term memory (LSTM) \cite{Hochreiter1997long}, and 3D convolutional neural networks (3D-CNN)\cite{shin2020convolutional} architectures, the proposed approach inherently accounts for motion information and can, therefore, also manage moving subjects. The latter is of particular interest since the proposed method can be helpful in typical real-world applications such as surveillance scenarios \cite{turaga2011diamond} by either supporting vision-based security systems or improving privacy concerns. In addition, a thorough search of the relevant literature yielded that Kato \textit{et al.} \cite{kato2021csi2image} are the only authors currently exploiting a generative modeling strategy based on wireless signals. However, their work focuses on the generation of still images. Furthermore, the model presented by Kato \textit{et al.}\cite{kato2021csi2image} is based on the traditional GAN structure supervised training scheme, and does not perform an actual radio-to-visual domain translation.

The effectiveness of the proposed method is evaluated by performing experiments on publicly available data capturing human poses of single individuals with commodity Wi-Fi in a constrained setting to ensure high-quality signals are received. Such an environment is important since Wi-Fi signals are absorbed by materials along their path and can be deformed by or superposed to other signals. For instance, when measuring a signal behind a reinforced concrete wall, it would result as greatly weakened with respect to an empty room \cite{adib2013see}. Indeed, preserving the signal quality is fundamental for Wi-Fi sensing applications, especially when a wireless communication could suffer severe interference from other radio devices or be completely denied through specific equipment such as Wi-Fi jammers \cite{hussain2014protocol}; and, as a matter of fact, constrained settings are typical for research-oriented and surveillance scenarios \cite{benito2020deeplerning}. In relation to the performed experiments, ablation studies on the pipeline were presented via quantitative performances, based on several state-of-the-art metrics for image quality evaluation between real and synthesized video frames. In addition, a qualitative assessment through visual observations, and further investigations on the method abstraction capabilities by replacing silhouettes in the training data with skeleton frames, demonstrated the effectiveness of the proposed approach to translate radio features into a visual domain representation. Interestingly, the synthesized videos satisfy and expand the concept of a privacy-conscious system \cite{lijie2020learning}; therefore, people may be monitored by perceiving and understanding their behaviors without collecting sensitive or private information, e.g., photos or audio recordings. 

Summarizing, the main contributions of this paper are as follows:
\begin{itemize}
    \item designing a novel two-branch generative Wi-Fi sensing framework that inherently considers motion information to synthesize coherent human silhouette and skeleton videos from wireless signals.
    \item describing a Wi-Fi data sanitization procedure through CSI measurements noise removal, which is key to obtaining robust radio features based on signals amplitudes, to increase the synthesized video quality.
    \item exploiting cross-modality supervision by organizing the 3D-GAN and LSTM models in a teacher-student fashion to transfer visual knowledge to Wi-Fi data, 
    achieving a privacy-conscious model that completely ignores video inputs without loss of generality on the synthesis task.
\end{itemize}

\section{Related Work}
Nowadays, intelligent computing and machine learning are the primary underlying technologies employed to develop automated algorithms and design learning strategies for finding the optimal solution to a large variety of tasks \cite{soroush2021game,jonghyun2021deep,avola2021lietome,yagiana2021improved,avola2020bodyprint}. Among these automated algorithms, generative adversarial learning has been widely used for image synthesis in several fields during the last few years. For instance, Isola \textit{et al.} \cite{isola2017image} address the paired image-to-image translation by investigating conditional GAN networks. They observed that conditioned generative modeling efficiently handles tasks requiring photographic output. Indeed, also Zhang \textit{et al.} \cite{zhang2017stackgan} use two stacked conditional GAN models for photo-realistic image synthesis from text data. However, they divide the text-to-image problem into two main stages. The Stage-I GAN generates low-resolution images while the Stage-II GAN, stacked on top of it, generates realistic high-resolution images conditioned by the previous stage results and text descriptions. Once again, GANs are also used by Krishna \textit{et al.} \cite{Krishna2019crossview} for cross-view image synthesis to produce, principally, outdoor scene images combining aerial and street image views. Sangkloy \textit{et al.} \cite{sangkloy2017scribbler} focus on the GAN-based sketch-to-image synthesis method, constraining the generative process with sketched boundaries and color strokes to obtain realistic images. Instead, Zhu \textit{et al.} \cite{zhu2017unpaired} use cycle-consistent adversarial networks (CycleGANs) for image-to-image translation but, differently from Isola \textit{et al.} \cite{isola2017image}, learn to translate an image from a source domain X to a target domain Y in the absence of paired training examples. Recently, the GAN is being increasingly used to synthesize human-related images. For instance, Yao \textit{et al.} \cite{yao2020gan} achieve remarkable results for person search in video sequences by combining the GAN capability with a deep complementary classifier based on a convolutional neural network (CNN). They used the GAN to generate new samples for the training set augmentation, improving the classifier performance during the testing phase. Fang \textit{et al.} \cite{fang2020identity}, instead, present an identity-aware CycleGAN for face photo-sketch synthesis and recognition. They improve the CycleGAN performance to solve the photo-sketch synthesis task by giving attention to the facial regions, including eyes and nose, which can be significant in identity recognition. Focusing on the dynamic synthesis of facial expressions, Gong \textit{et al.} \cite{gong2018dynamic} introduce a GAN working with a series of semantic parts with different shapes to describe geometrical facial movements. For person monitoring applications, instead, the GAN is primarily used to improve deep models performances \cite{zheng2017unlabeled,liu2019semantic,zhang2020pac-gan}. On a different note, a model trained on a specific dataset does not work when applied on another collection. Zhou \textit{et al.} \cite{zhou2019multicamera} solve such an issue by proposing a GAN-based image-to-image translation, transferring the images of a source dataset to the style of each camera in a target set of data. Similarly, Liu \textit{et al.} \cite{liu2019feature} propose a GAN structure to preserve features for cross-domain person re-identification, solving problems arising from significant variations between the training dataset and the target scene.

The state-of-the-art validates the incredible power of generative adversarial learning for image synthesis starting from text, sketch, or another image, as the information source. However, nowadays, the Wi-Fi signal is being explored for synthesizing visual data, opening up a new frontier for image synthesis and surveillance applications. Indeed, Wang \textit{et al.} \cite{wang2019person} propose a method for person perception using CSI measurements. They combine multiple residual convolution blocks and U-Net \cite{ronneberger2015unet} models to synthesize either human silhouette or pose. Similarly, only for the latter, Kefayati \textit{et al.} \cite{kefayati2020wi2vid} introduce a network architecture comprising three CNN-based sub-models: CSI encoder, domain translator, and frame decoder. The first encodes the CSI measurements, building the latent representations of radio features; the second connects such representations to the visual domain. Finally, the third generates images from the information in the latent visual domain. Again, Guo \textit{et al.} \cite{guo2020signal} propose a two-branch network to synthesize the human skeleton from a CSI-based image. The top branch leverages the OpenPose \cite{cao2021openpose} framework to extract, from video frames, the skeletal data used as supervision. Instead, the bottom branch generates the corresponding skeleton for each CSI image exploiting the supervision data as ground truth (GT). A similar supervision strategy is followed by Kato \textit{et al.} \cite{kato2021csi2image} for image synthesis of stationary subjects starting from the signal-to-noise ratio (SNR) of the Wi-Fi signal. They perform an ablation study on different classical GAN-based structures combined with a complementary detection system to quantitatively evaluate their synthesis capabilities. Despite the remarkable results, none of the tested methods considers the motion aspect of moving subjects. To address the latter, the introduced method improves the generative adversarial learning strategy by exploiting the 3D-GAN architecture. Moreover, the model generalization capabilities are further improved by manipulating the middle-level representations used to transfer visual knowledge to wireless signal-based features, ultimately enabling for the human silhouette and skeleton video synthesis via radio-to-visual domain translation.

\begin{figure*}
	\centering
	\includegraphics[width=\textwidth]{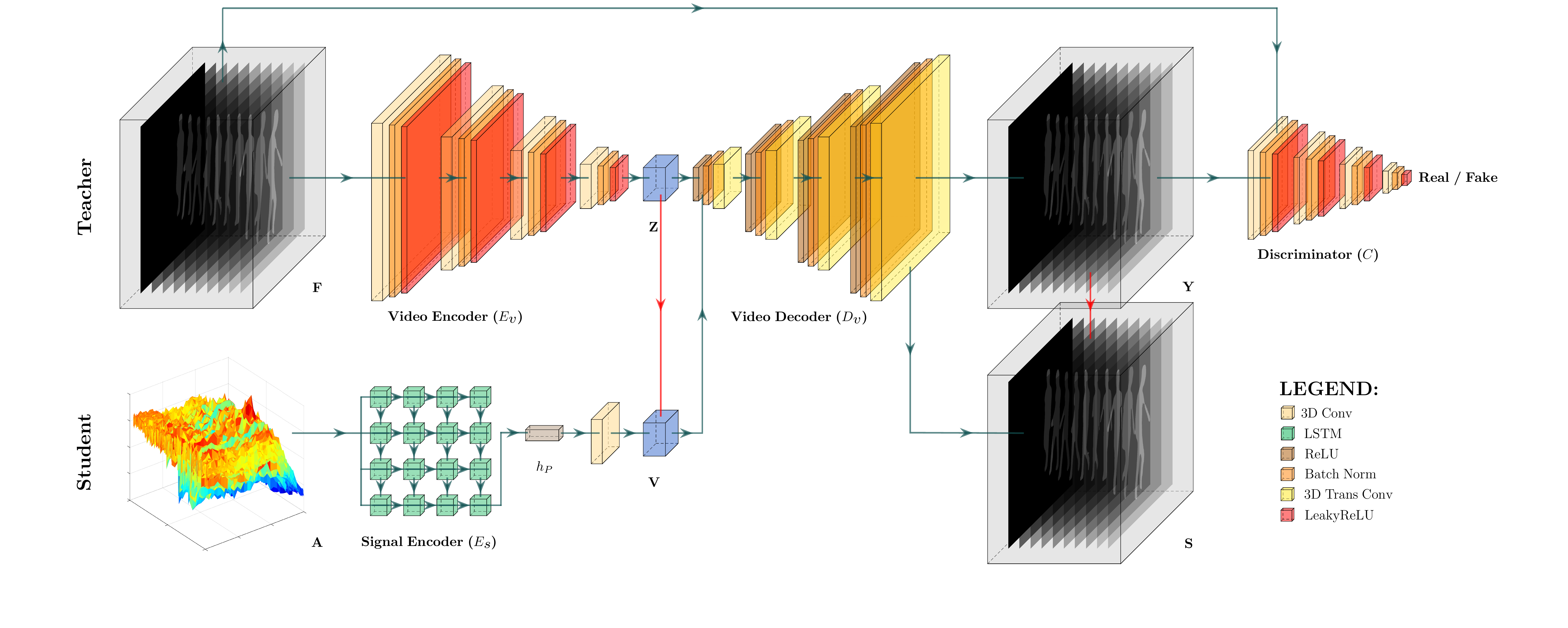}
	\caption{Proposed model architecture for video synthesis from Wi-Fi signals. Given a synchronized pair of human silhouette or skeleton video and CSI extracted amplitudes as input, visual knowledge is transferred to radio-based features by translating and mapping them in the visual domain via a teacher-student design. Red arrows indicate cross-modality supervision. Note that the student model can synthesize videos leveraging only Wi-Fi signals.}\label{fig:architecture}
\end{figure*}

Video synthesis is a direct consequence of image synthesis. It is concerned with the generation of new dynamic content rather than still images, enabling the perception and understanding of events in advanced automated systems, especially for surveillance and monitoring applications. To this end, Benito-Picazo \textit{et al.} \cite{benito2020deeplerning} propose a video surveillance system to detect anomalous moving objects by analyzing frames on the basis of probabilistic mixture distributions, exploited to generate windows within the video in proximity of areas where anomalies occur. Instead, Bang \textit{et al.} \cite{bang2021proactive} introduce a method to prevent accidents by monitoring construction sites through an unmanned aerial vehicle (UAV). Specifically, the UAV collected video frames are used to generate frames predicting future locations and posture of objects assisting the worker, including future speed and direction, effectively obtaining proactive safety information. Again for monitoring purposes, even Jaad \textit{et al.} \cite{jaad2020modeling} propose a future video frame generation-based system to predict the growth of urban areas during the years starting from historical images. Finally, unlike previous works, Cai \textit{et al.} \cite{enjiana2020selfadapted} focus on an optimization-based approach of self-adapted video magnification for subtle color and motion amplification that can be useful, for example, to estimate heartbeat by observing blood flow or other vital signs inside video sequences; once again highlighting how video synthesis might be crucial to solve diverse and complex tasks. 

\section{Proposed Method}\label{sec:proposed_method}
A novel two-branch neural network was designed and organized on parallel branches, sharing a decoder component to map radio signals to the visual domain and synthesize human silhouette and skeleton videos from Wi-Fi signals by emulating a teacher-student relationship. In particular, the teacher supervises the student training phase, transferring vision-based information to the associated sanitized amplitudes of the observed signal. The proposed architecture is summarized in Fig. \ref{fig:architecture}. Precisely, the teacher model is a 3D-GAN handling visual data that, after learning the low-dimensional manifold of observed videos about human silhouettes 
\begin{figurehere}
    \includegraphics[width=\columnwidth]{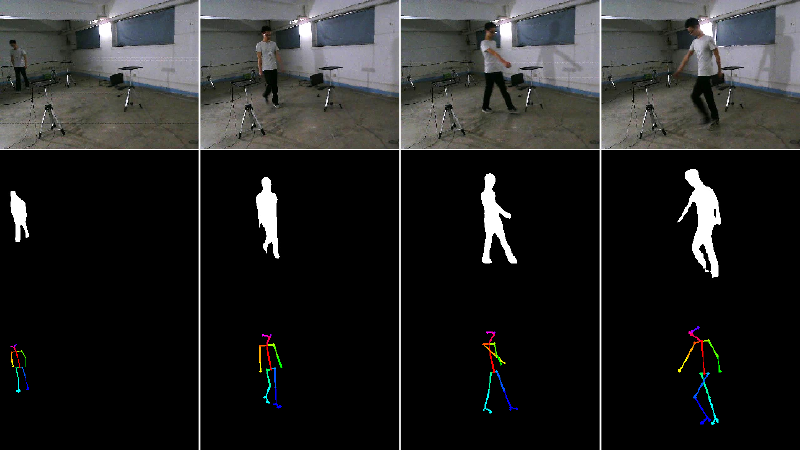}
    \caption{Video data input examples. The original RGB frames are reported in the top row, while the corresponding extracted human silhouette and skeleton, are shown in the middle and bottom rows, respectively.}
    \label{fig:silhouette_example}
\end{figurehere}
\noindent
or skeletons, produces data used as the visible ground truth for amplitudes processed by the student model. Such GAN type has been chosen since it has been proven effective in synthesizing both still images\cite{marriott20213d} and videos\cite{yan20183d}. The student is a novel hybrid autoencoder (AE) based on LSTM\cite{avola2020deep,avola2020lietome} and CNN\cite{avola2021ms,avola2021r} architectures, inspired by the domain translation devised by Zhu et al.\cite{zhu2017unpaired} for image-to-image synthesis and our previous experiences on training supervision\cite{avola2019master,avola2021multimodal}.
\begin{figure*}
	\centering
	\includegraphics[width=\textwidth]{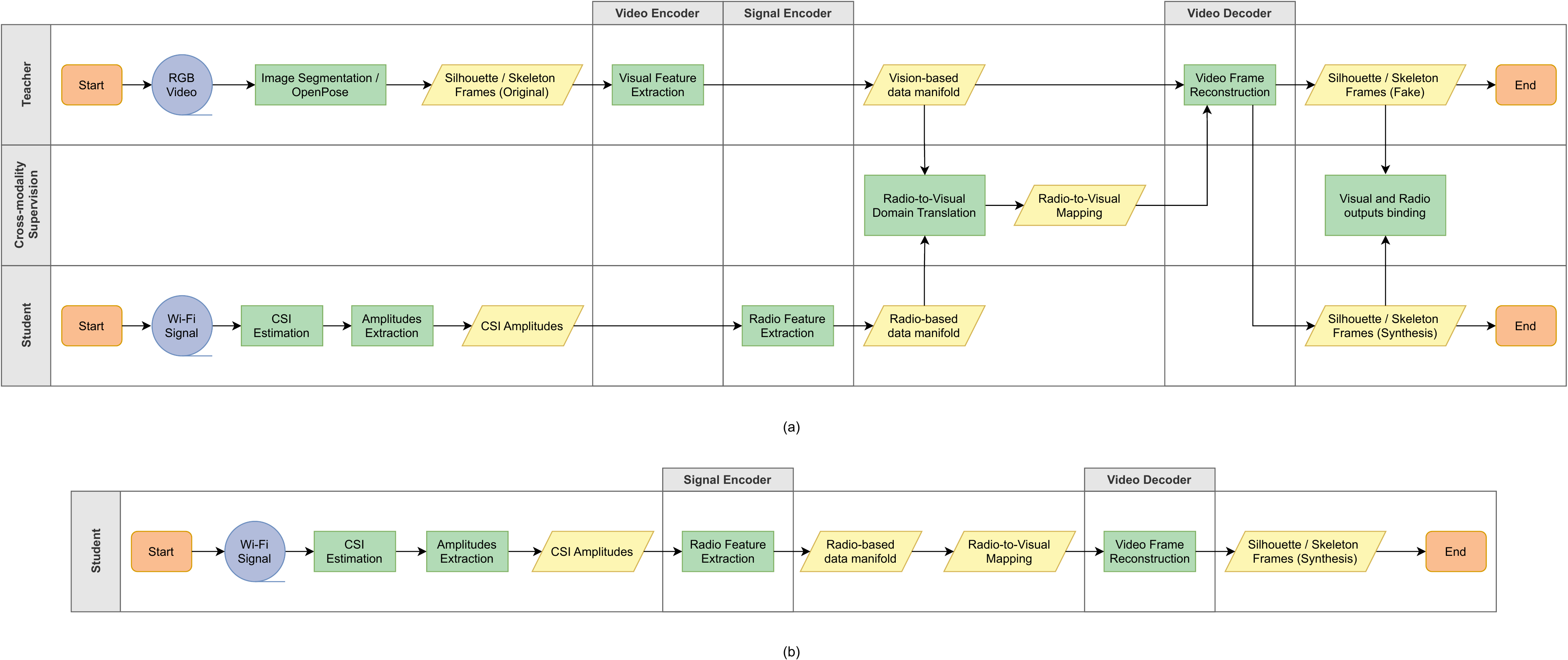}
	\caption{Proposed model workflows for video synthesis from Wi-Fi signals. In (a) and (b) the training and testing flowcharts, respectively.}\label{fig:flowchart}
\end{figure*}
In detail, the latter was specifically designed to handle amplitudes recorded over time, achieved by combining LSTM and 3D-CNN architectures to generate a latent radio representation of the signal. Then, by implementing a supervision from the teacher model, the student learns the effective mapping between radio and visual domains, i.e., translating amplitudes into silhouette or skeleton videos. Due to this distinctive design, the fundamental strategy and difference with existing works addressing image synthesis from Wi-Fi signals, is that the proposed network architecture exploits synchronized pairs composed by a human silhouette (or skeleton) video and CSI extracted amplitudes, both taken from the same underlying environment, of a person continuously performing different poses to synthesize accurate outputs. The human silhouette, shown in Fig.~\ref{fig:silhouette_example}, is obtained by applying the semantic image segmentation solution proposed by Chen \textit{et al.}\cite{chen2017rethinking} to an input RGB sequence. The latter is also used as a starting point to extract skeletons through the OpenPose\cite{cao2019openpose} framework. In this way, environment background and personal information are removed from the input, enabling the model to focus exclusively on the subject and its dynamics \cite{avola2019descriptor}, i.e., the person moving in the scene, like in most real camera-based surveillance scenarios \cite{avola2021machine}. Instead, the sanitized amplitudes are extracted from the CSI measurements of sequential Wi-Fi data packets as signal-based features describing human poses in the radio domain \cite{yongsen2019wifi}. This paired input enables the cross-modality supervision to learn a mapping from one domain to another during the network training phase. Accordingly, once the whole network is trained, only the student model and sanitized CSI extracted amplitudes are considered for the video synthesis. The result is a framework that can generate new person-related video frames from Wi-Fi signals without requiring any additional human or visual annotation as supervision as well as without any loss of generality. For illustration purposes, both training and testing workflows are depicted in Fig.~\ref{fig:flowchart}(a) and Fig.~\ref{fig:flowchart}(b), respectively.

\subsection{Channel State Information}
Regarding a standard wireless transmission, $P$ data packets characterize the Wi-Fi signal exchanged between fixed transmitting (TX) and receiving (RX) APs, integrating, respectively, $\Gamma > 1$ and $\Theta > 1$ antennas. In this multiple-input and multiple-output (MIMO) setting, the CSI is measured by employing the orthogonal frequency-division multiplexing (OFDM) transmission technology, including fine-grained signal information at the subcarrier level \cite{zheng2013rssi}. The CSI is a frequency-based measurement obtained by applying the fast Fourier transform (FFT) on the channel impulse response (CIR) at the receiver to compute the corresponding channel frequency response (CFR) \cite{shen2019wirim}. In practice, such a measurement estimates the CFR for each packet $p \in P$ reaching the RX device physical layer. Formally, in the frequency domain, the APs communication channel is linearly modeled as follows:
\begin{equation}\label{eq:wchannel}
    y = Hx + n,
\end{equation}
where $y$ is the vector of the received signal, $H$ is the CFR value, $x$ is the transmitted signal vector, and $n$ is the additive white Gaussian noise (AWGN) at reception. From Eq.~\eqref{eq:wchannel}, the OFDM technology provides a sampled CFR at the subcarrier level; thus, the CSI measurement is computed by including the CFR values from $K$ OFDM subcarriers defining the communication channel between RX and TX APs. Indeed, over the receiving $\theta \in \Theta$ and transmitting $\gamma \in \Gamma$ antennas, for each subcarrier $\kappa \in K$, the frequency response is a complex number that includes the signal amplitude $|H_{\kappa}^{(\theta,\gamma)}|$ and phase $\angle H_{\kappa}^{(\theta,\gamma)}$, and is defined as:
\begin{equation}
    	H_{\kappa}^{(\theta,\gamma)} = |H_{\kappa}^{(\theta,\gamma)}|e^{j \angle H_{\kappa}^{(\theta,\gamma)}},
\end{equation}
where $j$ represents the imaginary component of such a number. Finally, the CSI matrix obtained accounting all communicating antennas and $K$ subcarriers is the $\Theta \times \Gamma \times K$ matrix defined as:
\begin{equation}
    	CSI = \begin{bmatrix}
		H_{1}^{(1,1)} & H_{2}^{(1,1)} & \hdots & H_{\kappa}^{(1,1)}\\
		H_{1}^{(1,2)} & H_{2}^{(1,2)} & \hdots & H_{\kappa}^{(1,2)}\\
		\vdots & \vdots & \vdots & \vdots\\
		H_{1}^{(\theta,\gamma)} & H_{2}^{(\theta,\gamma)} & \hdots & H_{\kappa}^{(\theta,\gamma)}\\
	\end{bmatrix},
\end{equation}
where $H_{\kappa}^{(\theta,\gamma)}$ is the signed 8-bit complex CFR number for the $\kappa$-th subcarrier over the $\theta \in \Theta$ and $\gamma \in \Gamma$ antennas. According to the CSI specification, the amplitude can be derived from such a matrix but eventually requires further processing to be useful for Wi-Fi sensing applications.

\subsection{Signal Amplitude Sanitization}
To obtain meaningful radio-based features for synthesizing human silhouette videos, a procedure is designed to filter the CSI extracted amplitudes and mitigate noises due to wireless protocol specifications and environmental conditions, as can be observed in Fig. \ref{fig:amplitudes}(a). Indeed, abnormal values can appear in the CSI measurement and affect the extraction of human dynamics; therefore, such outliers should be removed \cite{kun2014pads}. Filtering the amplitude allows to suppress irrelevant radio information not necessarily correlated to human activity, i.e., mitigates noise caused by various factors such as furniture material and position or other external radio interference. To this end, it is possible to investigate the Hampel identifier \cite{davies1993identification} to recognize an outlier value as any point falling outside a closed interval $[\mu - \lambda \sigma,\mu + \lambda \sigma]$, where $\mu$ and

\begin{figurehere}
    \centering
    \includegraphics[width=\columnwidth]{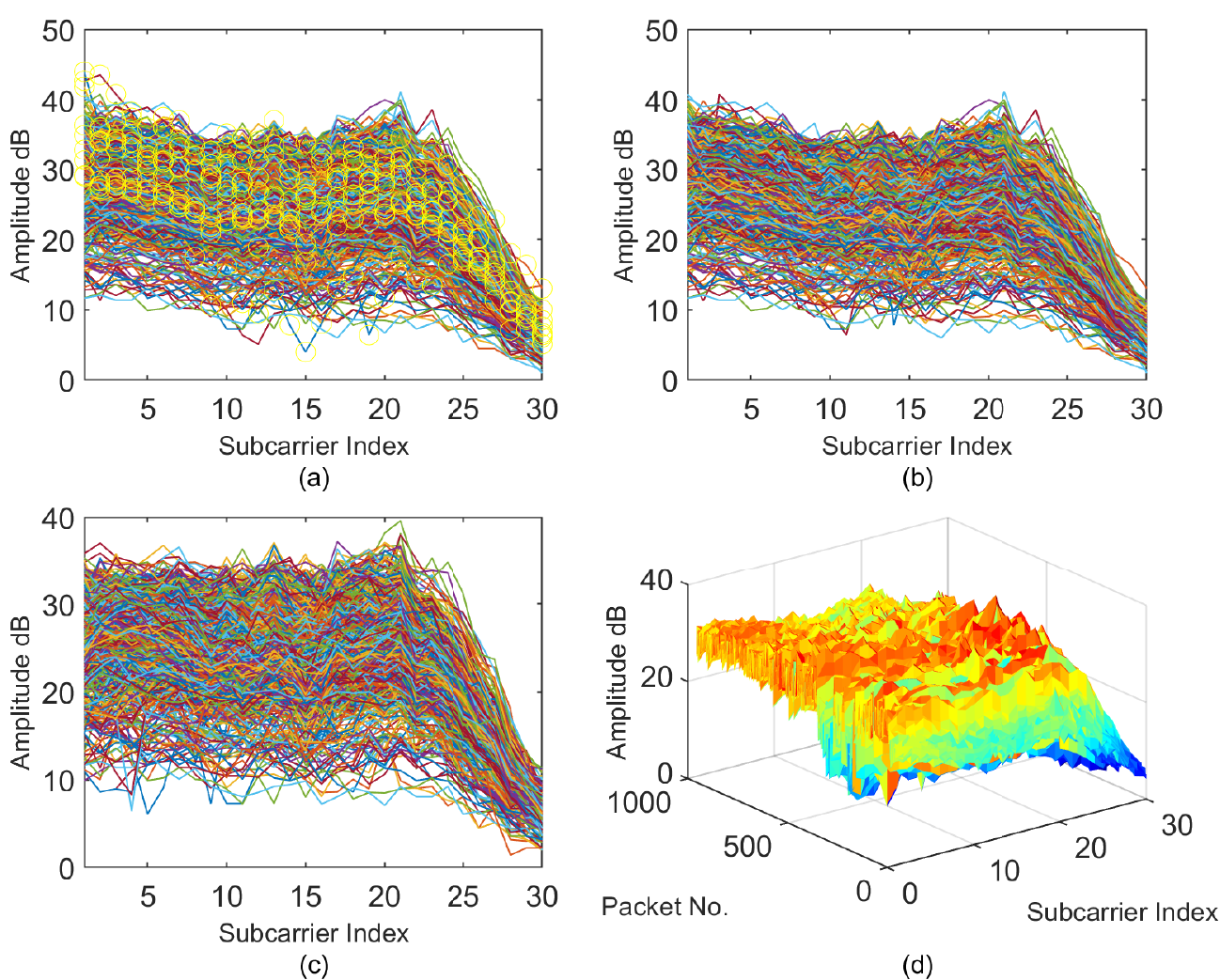}
    \caption{CSI extracted amplitudes processing example for $1000$ data packets. In (a) and (b) the raw and sanitized amplitudes for one TX-RX antenna pair, respectively. Yellow circles are abnormal values in raw data. In (c) and (d) the median filtering over the transmissions (i.e., all antenna pairs) and the corresponding 3D surface plot for a more comprehensive view.}
    \label{fig:amplitudes}
\end{figurehere}
\noindent
$\sigma$ are the median and median absolute deviation (MAD) of the data sequence, respectively, while $\lambda$ is an application-dependent constant. Precisely, given a sliding window of fixed length, the local outliers are detected by exploiting local median values over this window. Afterwards, the detected abnormal values are replaced with the previous non-outlier quantity to maintain consistent amplitude information. In detail, by empirically setting $\lambda = 3$, outliers are identified by points resulting in more than three local MAD away from the local median within the sliding window over the wireless data packets for each subcarrier. Formally, given the signal amplitudes extracted from the CSI measurements of $p \in P$ wireless packets transmitted between the TX and RX antennas, and considering the size of a window $\omega$, the local median is defined as follows:
\begin{equation}
   \mu(\Omega^{p,\kappa}) = \Omega^{p,\kappa}_{\lceil \omega/2\rceil},
\end{equation}
\begin{equation}
    \begin{split}
        \Omega^{p,\kappa} = \Big\{|H_{\kappa}|&^{p-\lfloor \omega/2\rfloor},\dots,|H_{\kappa}|^{p+\lfloor \omega/2\rfloor}:\\
        &\ |H_{\kappa}|^{p-\lfloor \omega/2\rfloor}  < |H_{\kappa}|^{p+\lfloor \omega/2\rfloor} \Big\},
    \end{split}
\end{equation}
where $\Omega^{p,\kappa}$ is the set containing $\omega$ neighboring packets amplitude $|H_{\kappa}|$ of the $\kappa$-th subcarrier, in ascending order. Notice that the equation of a single sample is reported for the sake of simplicity; however, $\mu$ is computed over all $\Theta \times \Gamma \times K$ antennas and subcarriers combinations. Therefore, the local MAD used to detect abnormal amplitude values is defined as:
\begin{equation}
    \begin{split}
    \sigma(\Omega^{p,\kappa})& =\ \mu(|\Omega^{p,\kappa}_i - \mu(\Omega^{p,\kappa})|), \\
    & \forall i,\ \text{s.t.}\ 1 \leq i \leq \omega.
    % &\forall i \in [0, w].
    \end{split}
\end{equation}
Finally, the intervals in which points are acceptable local values are defined as:
\begin{equation}
    limit^{p,\kappa} = \mu(\Omega^{p,\kappa}) \pm 3 * \sigma(\Omega^{p,\kappa}),
\end{equation}
\noindent 
and each value falling outside these ranges is replaced with the previous non-outlier value to maintain information consistency. The produced sanitized amplitudes, for an empirically chosen window size $\omega = 50$, are shown in Fig. \ref{fig:amplitudes}(b). Since the sanitized $\Theta \times \Gamma$ transmissions share similar amplitudes properties, median filtering is applied over the transmissions to condensate such properties in a matrix $A$ with size $P \times K$, as shown in Fig. \ref{fig:amplitudes}(c) and Fig. \ref{fig:amplitudes}(d). In this way, data dimensionality is reduced, and amplitudes shared among different antennas transmissions are concentrated. To train the two-branch network, the CSI extracted amplitudes are paired with the synchronized videos used to supervise the synthesis process.

\subsection{Two-branch Network Architecture}
Starting from the sanitized amplitudes paired with the corresponding synchronized video, the mapping between radio and visual features is achieved by training the two-branch network in a teacher-student fashion. In detail, to find this mapping, the network exploits two parallel branches sharing the same decoder component. The top branch has a 3D-GAN structure handling vision-based data and acts as the teacher model. Specifically, the latter consists of video frames encoder $E_v$, decoder $D_v$, and discriminator $C$ components. In particular, for the $E_v$ and $C$ models, there are $3$ 3D convolutional layers, each comprising the sequence of strided 3D convolution, batch normalization \cite{ioffe2015batch}, and leaky rectified linear unit (leakyReLU) \cite{maas2013rectifier} activation function. However, these two components differ in the last layer. In fact, being $C$ a binary classifier, it applies a sigmoid activation function to the last 3D convolution output. Regarding the decoder $D_v$, it follows a reverse structure with respect to the encoder $E_v$ with $3$ 3D transposed convolutions, rather than convolutions, and implements ReLU instead of LeakyReLU activation functions to stabilize the training process, as suggested by Radford \textit{et al.} \cite{radford2016unsupervised}. Moreover, after the last transposed convolution, $D_v$ uses the hyperbolic tangent function to reconstruct video frames. Notice that strided 3D convolutions are used rather than traditional 2D-based ones, which are generally followed by a max-pooling operation, and the reason is twofold. First, the stride reduces the computational cost and dynamically learns the pooling operation, improving the entire model generalization \cite{springenberg2015striving}. Second, the 3D convolution effectively performs video analysis capturing both spatial and temporal information. Indeed, the 3D-GAN goal is to learn how to reproduce the observed human silhouette, or skeleton, videos distribution, leveraging a low-dimensional manifold that comprises feature maps in visual and temporal domains to keep track of human poses across the video frames. Intuitively, given a sequence of $L$ 3D convolutional layers, each of them extracts spatial and temporal characteristics from the local neighborhood on the feature maps connected to various frames in the corresponding previous layer. Subsequently, a bias is applied, and an activation function is used on the result to generate feature maps on the current layer. The temporal dimension is caught convolving the 3D kernel on stacked contiguous video frames, allowing for the extraction of motion information from the video. Formally, for each feature map $j \in J_l$ computed in layer $l \in L$, the 3D value $v$ at position $(x,y,z)$ in $j$ is defined as:
\begin{equation}
    \begin{split}
         \mathop{v_{lj}^{xyz}} = \phi\bigg(b_{lj}+ \sum_m & \sum_{w=0}^{W_l -1} \sum_{h=0}^{H_l -1} \sum_{d=0}^{T_l -1} \\ &\mathop{\rho_{ljm}^{wht}} \mathop{v_{(l-1)m}^{(x+w)(y+h)(z+t)}}\bigg),
    \end{split}
\end{equation}
where $\phi$ is the activation function; $b_{lj}$ describes the bias for the current feature map; $m$ indicates the feature map index of the previous layer $(l-1)$ connected to the $j$-th feature map; $W_l$, $H_l$, and $T_l$ correspond to the height, width, and temporal depth of the 3D kernel, respectively; while $\mathop{\rho_{ljm}^{wht}}$ represents the the kernel $\mathop{\rho_{ljm}}$ value at position $(w, h, t)$ connected to the $m$-th feature map. This characterization allows the teacher model to learn the latent space $Z$, with size $J_L \times T_L \times H_L \times W_L$, representing multiple contiguous frames with $J_L$ feature maps of size $T_L \times H_L \times W_L$. Observe that this space includes low-dimensional spatial and temporal features describing human silhouette, or skeleton, poses associated to the visual domain.

Concerning the bottom branch of the proposed network, it has a hybrid AE structure handling radio-based data and acts as the student model. Precisely, the latter comprises an LSTM-based signal encoder $E_s$ and shares the video frames decoder $D_v$ of the teacher. This type of architecture is significant for this branch because it effectively learns to map radio features to vision-based data. In particular, for the encoder $E_s$, there is a LSTM layer with $P$ units, i.e., one per packet, and a 3D transposed convolution is applied on the last unit result to enable the radio-to-visual domain translation. The LSTM was chosen since it can extract features from sequential data \cite{xiuhui2020human}. In fact, this architecture has proven to be an ideal solution to learn the low-dimensional radio features from the sequence of CSI extracted amplitudes of contiguous wireless data packets. Afterwards, such features are translated and employed to synthesize video frames through the $D_v$ component. Note that each LSTM unit retains important features computed from the amplitude sequence by exploiting its input, forget, and output gates to update a cell state; allowing the model to forget otherwise irrelevant information\cite{greff2017lstm}. Formally, given a sequence of CSI amplitudes $A = \{a_{1,1}, a_{1,2}, \dots, a_{P,K} \}$, over the subcarriers $\kappa \in K$, for each packet $p \in P$ the corresponding $\text{LSTM}_{p}$ unit is defined as:
\begin{equation}
    \begin{split}
        & i_{p} = \sigma (\Pi_{ai} a_p + U_{hi} h_{p-1} + \Psi_{ci} c_{p-1} + b_i), \\
        & f_{p} = \sigma (\Pi_{af} a_p + U_{hf} h_{p-1} + \Psi_{cf} c_{p-1} + b_f), \\
        & o_{p} = \sigma (\Pi_{ao} a_p + U_{ho} h_{p-1} + \Psi_{co} c_{p-1} + b_o), \\
        & \Tilde{c}_{p} = tanh(\Pi_{a\Tilde{c}} a_p + U_{h\Tilde{c}}\ h_{p-1} + b_{\Tilde{c}}), \\
        & c_{p} = f_{p} \odot c_{p-1} + i_{p} \odot \Tilde{c}_{p}, \\
        & h_{p} = o_{p} \odot tanh(c_{p}), \\
    \end{split}
\end{equation}
where $i$, $f$, and $o$ indicate the the input gate, forget gate, output gate; $h$, $\Tilde{c}$, and $c$ denote the hidden state, cell update, and cell state, respectively; $\Pi$, $U$, and $\Psi$ are the weight matrices for the corresponding gates, hidden states, and peep-hole connections; while $b$ indicates a bias vector added to every gate or cell update. Therefore, the low-dimensional radio features learned by $E_s$ for $P$ packets are represented by the last LSTM unit hidden state vector $h_P$; capturing an abstract representation of the whole input sequence. Upon extracting these radio features, the 3D transposed convolution is applied to prepare the $h_p$ vector for radio-to-visual domain translation, obtaining a new latent representation $V$ that enables the silhouette, or skeleton, video frames synthesis through the $D_v$ component.

\subsection{Cross-modality Supervision}
During the training phase, $N$ pairs of synchronized data $<F,A>_n$ are used as the model input, where $F$ and $A$ correspond to the set of human silhouette, or skeleton, frames and sanitized CSI amplitudes extracted from the Wi-Fi signal associated to the video, respectively, for the $n$-th dataset sample. In detail, for each sample $n \in N$, the encoder $E_v$ takes as input the set $F$ containing real frames and computes the corresponding latent space $Z$. Afterwards, this low-dimensional representation is used by the decoder $D_v$ to reconstruct the original video, defining a set of fake frames $Y$. On the other hand, the encoder $E_s$, supervised by the teacher branch, takes as input the sanitized amplitudes $A$ and computes the latent vector $h_P$ which, analyzed through the 3D transposed convolutional operation, outputs a latent space $V$, with the same shape of $Z$, that is used for the radio-to-visual translation. Finally, the shared component $D_v$ synthesizes video frames $S$, corresponding to the original input set $F$, using latent space $V$. In general, mapping data from radio to visible spectrum is challenging due to the lack of labeled data. This problem is solved by employing the teacher branch to generate ground truth data leading to the domain-to-domain translation. In this proposed teacher-student design, the vision-based information is transferred to signal features by linking the latent representations of video and CSI extracted amplitudes.  

Regarding the teacher model, the 3D-GAN learns a latent manifold of input videos required for the reconstruction goal, employing the classical GAN adversarial function based on the zero-sum game. Formally, this is achieved through the following objective loss derived from the cross-entropy between real and fake videos, defined as:
\begin{equation}
    \begin{split}
        &Z = E_v(F), \\
        &Y = D_v(Z), \\
        &\mathop{\mathcal{L}_F} = E_{F} [log\ C(F)],\\
        &\mathop{\mathcal{L}_{Y}} = E_{Y} [log\ (1-C(Y)],\\
        &\mathop{\mathcal{L}_{adv}}  = \min_{\{E_v,D_v\}}\max_{C}\ ( \mathop{\mathcal{L}_F} + \mathop{\mathcal{L}_{Y}}),
    \end{split}
\end{equation}
where $Z$ and $Y$ are the latent space and reconstructed set of fake video frames; $C(\cdot)$ indicates the discriminator estimated probability of either real or fake videos being effectively real; while $E_F$ and $E_Y$ are the expected value over all the original and fake sets of frames. The proposed teacher network, generating $Y$ from the low-dimensional features, tries to reproduce the real video $F$ given as input. Therefore, the mean squared error is computed between video frames of $F$ and $Y$, as follows:
\begin{equation}
    MSE_Y = \frac{1}{T} \sum_{i=1}^T{(F_i - Y_i)^2},
\end{equation}
where $T$ indicates the input video number of frames. Finally, the training objective for the 3D-GAN is computed via the following weighted equation:
\begin{equation}
    \underset{Teacher}{\mathop{\mathcal{L}}} =  w_{adv}\ \mathop{\mathcal{L}_{adv}} + w_Y\ MSE_Y,
\end{equation}
where $w_{adv}$ and $w_Y$, with $w_{adv} < w_Y$, are weights adjusting the impact of each objective to the overall function. Notice that the obtained latent space $Z$ and reconstructed set of video frames $Y$ are key elements that enable cross-modality supervision for the student model.

Concerning the student model, it is implemented through a hybrid AE network that learns a low-dimensional representation of radio-based features by analyzing sanitized amplitudes via its $E_s$ module. In addition, due to the supervision process, the student can also find a feature mapping between the visual and radio domains. Formally, given the CSI extracted amplitudes $A$ for $P$ contiguous wireless data packets, the latent vector $h_P$ is computed as follows:
\begin{equation}
    h_P = E_s(A).
\end{equation}
Afterwards, a 3D transposed convolution is applied on this latent representation to obtain low-dimensional feature maps $V$ with the same shape of $Z$. Subsequently, the radio-to-visual domain translation is obtained by binding the two latent spaces $Z$ and $V$, effectively transferring knowledge from the visual domain, i.e., $Z$, into the radio one, i.e., $V$. Formally, this can be achieved by defining the following objective function for the student encoder $E_s$:
\begin{equation}
    % MSE_V = \frac{1}{2T} \sum_{i=1}^T{(Z_i - V_i)^2},
    MSE_V = \frac{1}{|Z|} \sum_{i=1}^{|Z|}{(Z_i - V_i)^2},
\end{equation}
where $|Z|$ correponds to the latent space size. Moreover, to further improve the student abstraction capabilities, the reconstructed frames $S$, generated from the radio latent space $V$ through the shared decoder $D_v$, are constrained to the set of frames $Y$ produced by the teacher. This allows to correctly transfer knowledge by binding the two outputs, i.e., $Y$ and $S$, and, consequently, to generate frames that are more similar to the real video frames $F$ independently from the exploited input type, i.e., video or radio. Formally, this second constraint is defined as:
\begin{equation}
    MSE_S = \frac{1}{T} \sum_{i=1}^T{(Y_i - S_i)^2},
\end{equation}
where $T$ corresponds, again, to the input video number of frames. Notice that the two objective functions $MSE_V$ and $MSE_S$ are constraints required for the domain-to-domain translation between video and signal amplitudes pairs. In particular, latent space $Z$ and reconstructed frames $Y$ act as ground truth data enabling cross-modality supervision from the teacher during the whole network training process. Finally, the training objective for the hybrid AE, i.e., student model, is computed through the following equation:
\begin{equation}
    \underset{Student}{\mathop{\mathcal{L}}} =  w_{V}\ MSE_V + w_{S}\ MSE_S,
\end{equation}
where $w_V$ and $w_S$, with $w_V < w_S$, are weights adjusting the impact of each objective to the overall function. Concluding, the entire two-branch network objective is to minimize the following loss function:
\begin{equation}
    \mathop{\mathcal{L}} = \underset{Teacher}{\mathop{\mathcal{L}}} + \underset{Student}{\mathop{\mathcal{L}}}
\end{equation}

\section{Experimental Results}
This section first describes the public dataset used to evaluate the proposed architecture, which is focused on capturing human poses of single persons with commodity Wi-Fi. Then it provides implementation details, including the chosen hyperparameters and employed hardware. Finally, quantitative and qualitative results are reported on the public collection mentioned above to present a comprehensive evaluation of the two-branch network. Observe that, although the teacher branch is fundamental for cross-modality supervision, it is exclusively utilized during the training phase. Instead, at evaluation, the student is detached from the other branch and tested directly by using sanitized signal amplitudes as input and by comparing its reconstructed video with the real input frames paired with the Wi-Fi signals. Moreover, to further investigate the proposed cross-modality supervision strategy effectiveness in finding 

\begin{figurehere}
    \centering
    \includegraphics[width=.6\columnwidth]{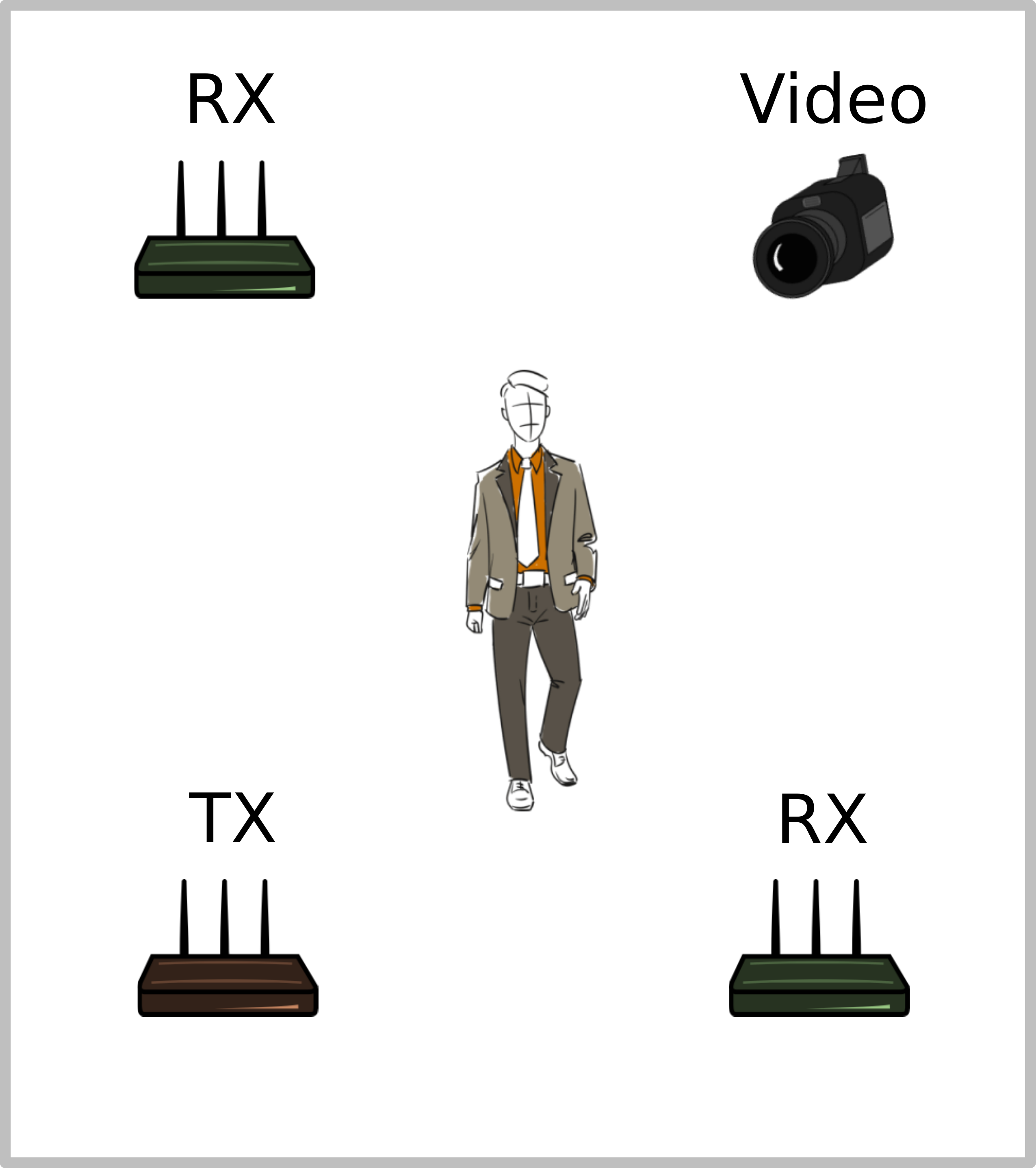}
    \caption{
    %\color{red}
    Transmitter (TX), receivers (RX), and RGB camera locations used for data collection in the experimental setting proposed by Guo \textit{et al.}\cite{guo2020signal}.}\label{fig:dataset_protocol}
\end{figurehere}
\noindent
a mapping between radio features and vision-based human representations, experiments were performed by reconstructing either silhouette or skeleton videos exploiting exclusively radio signals as inputs.

For the image quality assessment, three state-of-the-art metrics are reported concerning the quantitative results, i.e., the mean squared error (MSE), structural similarity index (SSIM) \cite{zhou2004image}, and feature similarity index (FSIM) \cite{zhang2011fsim}. Note that, even though the MSE is considered as the traditional measurement, it only considers pixel-by-pixel intensity comparison between original and synthesized video frames, therefore ignoring image structures. 
Instead, the SSIM and FSSIM address this issue by considering the structural and feature similarity, respectively.

\subsection{Dataset}
The publicly available data collection by Guo \textit{et al.} \cite{guo2020signal} contains $4.420$ video frames in total, depicting human subjects freely performing poses in a $7m \times 8m$ room. Furthermore, each video is associated with wireless data counting $1.000$ CSI samples for each RX antenna. In detail, the CSI was measured using the CSI Tool by Halperin \textit{et al.} \cite{halperin2011tool}, and  Wi-Fi signals were acquired using $3$ transceivers divided into $1$ transmitter and $2$ receivers, working in a $5$GHz frequency band with $20$MHz channel bandwidth. The former includes $\Gamma = 3$ TX antennas, while the latter each have $\Theta = 3$ RX antennas. What is more, as shown in Fig. \ref{fig:dataset_protocol}, the receivers were placed perpendicularly to one another to increase the wireless signals resolution. As a matter of fact, according to the Fresnel zone model \cite{wu2017devicefree}, two transceivers cannot capture a person walking parallel to the Line of Sight (LoS) path. In addition, an RGB camera was attached to a receiver to allow for synchronized video recordings and CSI samples. In particular, receivers were synchronized via the network time protocol (NTP), while wireless and video data, corresponding to a sample pair, were synchronized utilizing timestamps, with an average error of less than $1.5$ms. Finally, to adapt this collection for the video synthesis task, the algorithms mentioned in Sec.~\ref{sec:proposed_method} were employed to generate human silhouette and skeleton videos from the RGB sequences, i.e., semantic image segmentation and OpenPose, respectively.

\subsection{Implementation Details}\label{sec:implementation}
The proposed two-branch architectural design has been implemented using the Pytorch framework, and the Wi-Fi signals were processed via the MATLAB R2021a software. To correctly evaluate the proposed approach, the same protocol devised by the dataset authors Guo \textit{et al.} \cite{guo2020signal} is used for all the tests. Specifically, $75\%$ of the data was used to train the network, and the remaining $25\%$ for tests. Furthermore, each network was trained for 800 epochs using the Adam optimizer \cite{diederik2015adam} with an initial learning rate set to $0.0002$, an $\epsilon$ numerical stability parameter of $1e$-8, first momentum term $\mathcal{B}1$ with value $0.5$ and, finally, second momentum term $\mathcal{B}2$ with value $0.999$. Moreover, model weights were initialized from a zero-centered Gaussian distribution with standard deviation $0.02$, and the LeakyReLU slope in $E_v$ and $C$ components is set to $0.2$. Observe that these settings are suggested for stabilizing the GAN-based networks training phase \cite{radford2016unsupervised}. Regarding the weight parameters employed to adjust teacher and student models losses, they were empirically set to $0.5$ for $w_{adv}$ and $w_V$, and to $1$ for $w_Y$ and $w_S$. Finally, all reported experiments were performed using a single GPU, i.e., a GeForce RTX 2080 with 16GB of RAM. Despite the research-oriented scenario, the measured running time for the test phase suggests that the proposed method can be suitable for real-time constrained applications. In fact, once trained, the student model takes only $\approx$2.4ms on the GeForce GPU to synthesize each silhouette or skeleton video frame from the wireless signal. Therefore, to further investigate running time, the test phase was also executed 
\begin{figurehere}
    \includegraphics[width=\columnwidth]{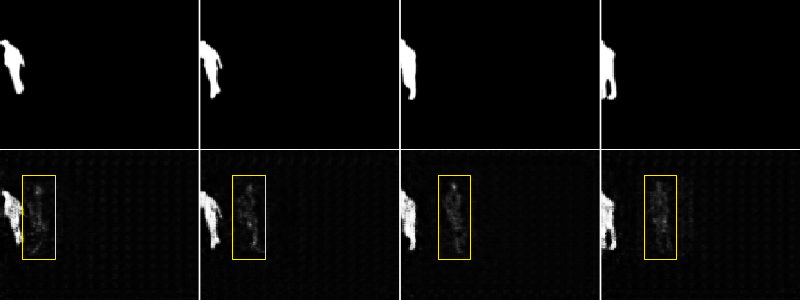}
    \caption{Examples of synthesized human silhouette for $h_P$ with size of $100$. In the top row, the silhouette frames representing the ground truth. In the bottom row, the noisy synthesized silhouette affected by the ghost effect, identified through a yellow rectangle.}
    \label{fig:ghost}
\end{figurehere}
\noindent
on a Raspberry Pi4 (RPi4) Model B with 8GB of RAM through the ONNX Runtime to simulate a possible low-cost real environment checkpoint for people activities monitoring through synthesized visual data. On the RPi4 board, the trained student model requires $\approx$32.6ms to synthesize a single frame, generating roughly 30 frames per second. This last experiment confirmed the feasibility of real-time usage in constrained surveillance scenarios. For completeness, the training time on the GeForce GPU took $\approx$5 and $\approx$13 hours to learn the synthesis of human silhouettes and skeletons, respectively. The training time differs according to the visual knowledge complexity employed for the radio-to-visual mapping.

\subsection{Silhouette Synthesis Evaluation}
The first batch of experiments evaluates the architecture capabilities to reconstruct human silhouette videos starting from Wi-Fi signals. The proposed solution allows the student model to learn radio-to-visual features translation which can focus exclusively on human body dynamics even though Wi-Fi signals contain coupled scattering patterns of the human body and environment. As a matter of fact, by retaining knowledge from the teacher, it is possible to transform CSI extracted amplitudes into features discriminating human-related information. More precisely, unlike existing methods that directly tune the network output, the proposed approach acts on the low-dimensional radio features latent space representation $V$ associated with the visual domain $Z$ via cross-modality supervision. Therefore, since $V$ and $Z$ have the same shape due to the architecture structure, vector $h_P$ size results critical for the domain-to-domain translation as it regulates the amount of information extracted from radio signals.

\begin{tablehere}
	\tbl{Latent signal-based feature vector $h_P$ size performance evaluation for human silhouette video synthesis. \label{tab:silhouette}}
	{\begin{tabular}{@{}cccc@{}}
		\toprule
		$h_P$ size & $MSE\downarrow$ & $SSIM\uparrow$ & $FSIM\uparrow$ \\ \colrule
		100 & 0.007 & 0.781 & 0.972 \\
		200 & 0.035 & 0.865 & 0.970 \\
		\textbf{300} & \textbf{0.002} & \textbf{0.885} & \textbf{0.990}  \\
		400 & 0.002 & 0.828 & 0.984 \\
		\botrule
	\end{tabular}}
\end{tablehere}
\noindent
Accordingly, tests were performed to evaluate the translated latent space $V$ effectiveness by changing the size of $h_P$, i.e., radio-based feature vector, ranging from $100$ to $400$ elements. The quantitative evaluation for this ablation study is reported in Table \ref{tab:silhouette}. As can be observed, the MSE measurement is close to 
\begin{figure*}
    \centering
    \includegraphics[width=\textwidth]{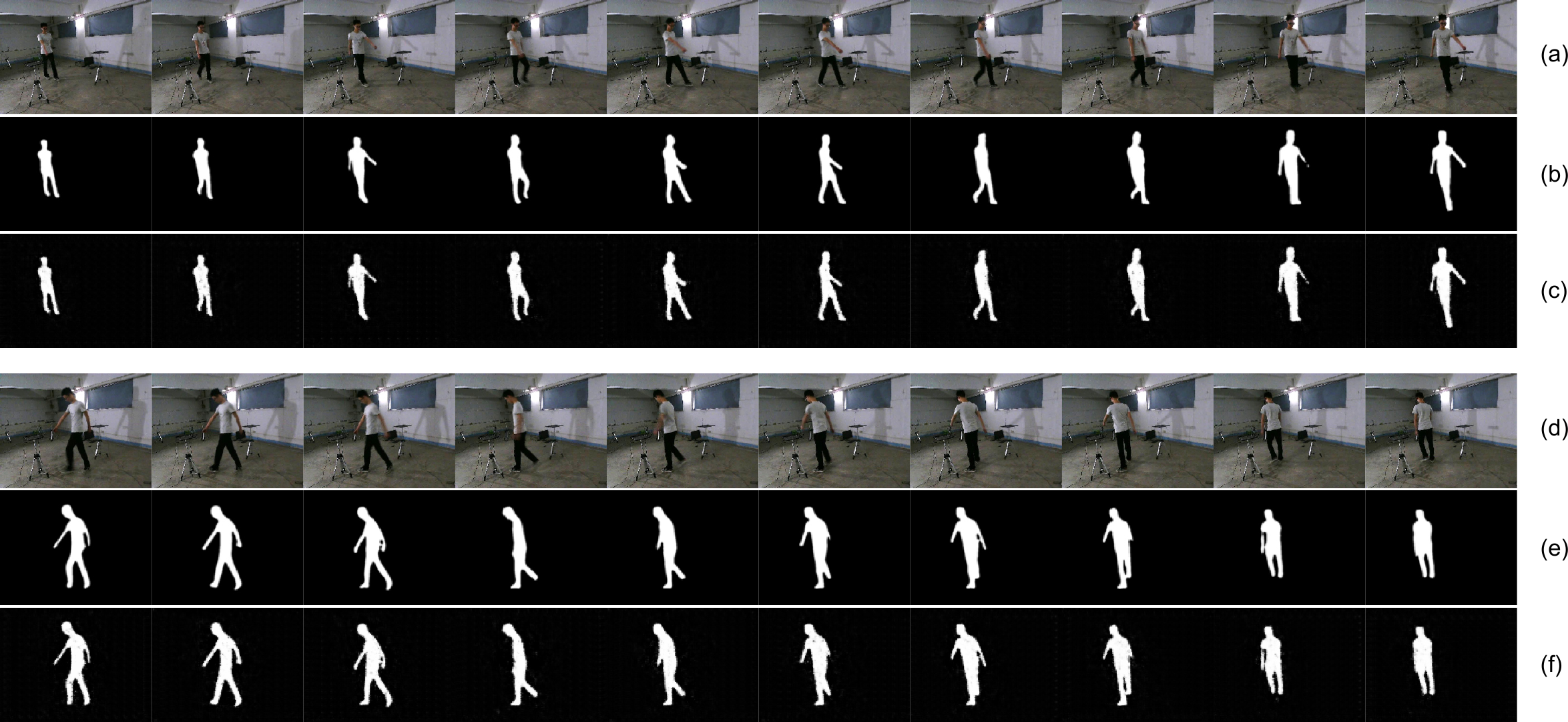}
    \caption{Test samples showing the human silhouette video synthesis for $h_P$ vector with the size of 100. In (a) and (d), the original RGB video frames are reported as visual reference. In (b) and (e), silhouettes extracted from RGB frames representing the ground truth. Finally, in (c) and (f), the silhouettes synthesized exploiting exclusively Wi-Fi signals.}
    \label{fig:results_silhouette}
\end{figure*}
\noindent
zero for all $h_P$ sizes, meaning that the student model can reconstruct accurate silhouette videos pixel-wise. The high FSIM scores also confirm the latter. In fact, this measure indicates a high image quality with respect to the expected output, i.e., ground truth silhouette videos. Significant performances are also achieved through the SSIM metric, which corresponds to structural similarities between the GT and generated output. However, by using a small $h_P$ size, i.e., $100$, the extracted signal features cannot correctly describe human movements in a scene, resulting in a noisy silhouette. This effect can be observed in Fig.~\ref{fig:ghost}, where synthesized frames present after images in the form of ghost silhouettes performing random poses due to the low representation capability derived from the small $h_p$ size. What is more, by increasing the radio feature size, i.e., $|h_p|=400$, performances start to decrease due to the extracted features capturing other background information. Consequently, for human silhouette generation, the best $h_p$ size is a vector with dimension $300$. Synthesized images for this configuration are shown in Fig.~\ref{fig:results_silhouette}, where the generated silhouettes are extremely similar and coherent with the expected output.

The human silhouette synthesis evaluation is concluded by presenting a qualitative comparison with the work introduced by Wang \textit{et al.}\cite{wang2019person} that, according to a thorough search of the relevant literature, is the only one performing silhouette generation from Wi-Fi signals to achieve person perception. Notice that a quantitative comparison cannot be reported since Wang \textit{et al.} performed their experiments on a private collection and did not employ standard reconstruction metrics, such as MSE, but instead implemented a segmentation measure to evaluate their method specifically. Regardless, a qualitative comparison, albeit carried out on different images, is presented in Fig.~\ref{fig:comparison_shape}. As can be observed, silhouettes synthesized by the presented approach, using radio features with size $|h_p|=300$, have higher image quality and show more detailed silhouettes. Such an outcome highlights the cross-modality supervision effectiveness in this domain-to-domain translation, which is achieved by mapping Wi-Fi signals to a visual domain through the knowledge transferred from the teacher model to the student one at training time.

\subsection{Skeleton Synthesis Evaluation}

In this second group of experiments, to further assess the proposed cross-modality supervision strategy effectiveness, the architecture is evaluated by replacing silhouettes in the video-radio signal training pairs with skeleton videos as an alternative vision-based information. Human skeletons, obtained by applying the OpenPose framework on RGB videos, were chosen since they are one of the most widespread human body representations\cite{avola20202d}. To train the architecture on human
\begin{figurehere}
    \includegraphics[width=\columnwidth]{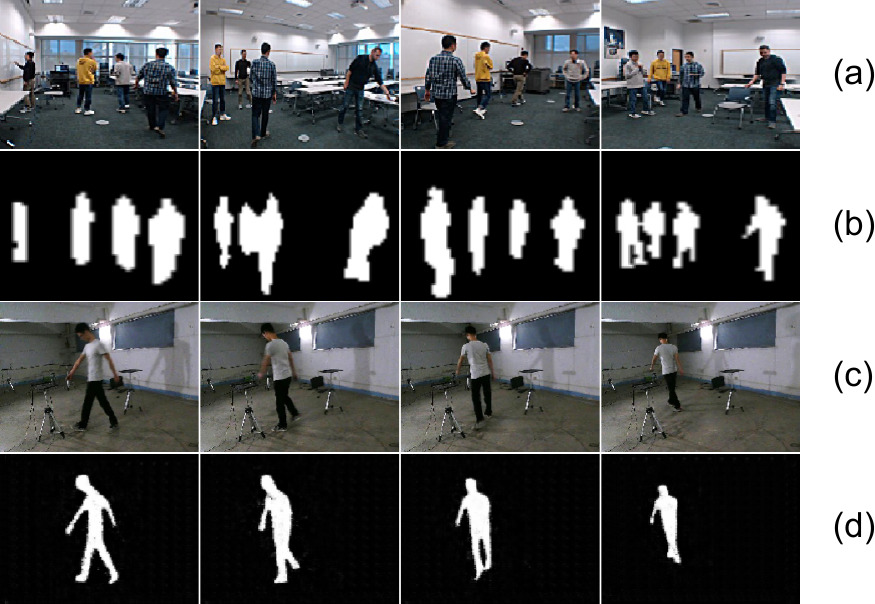}
    \caption{Qualitative comparison for the human silhouette synthesis. In (a) and (c), RGB visual references. In (b) and (d), the human silhouette synthesized from Wi-Fi signals by Wang \textit{et al.}\cite{wang2019person} and the proposed method, respectively.}
    \label{fig:comparison_shape}
\end{figurehere}
\noindent
skeleton synthesis from Wi-Fi signals, the same implementation details described in Sec. \ref{sec:implementation} were employed. However, the system required to be trained for $1600$ epochs to obtain high-quality images due to the fine-grained skeleton representation of OpenPose. Moreover, as mentioned in the previous section, since $h_p$ regulates the amount of information used in the latent representation $V$ used for radio-to-visual translation, the same tests on $h_P$ vector size were performed, ranging from $100$ to $400$ elements. The quantitative results for the student model are summarized in Table \ref{tab:skeleton}. As shown, all $h_p$ sizes provide roughly the same performance in feature similarity and pixel intensity comparisons, i.e., FSIM and MSE metrics, respectively. Concerning the structural similarity measure, i.e., SSIM, instead, it can be noticed that sizes $100$ and $400$ have higher performances in comparison with sizes $200$ and $300$. This outcome has a twofold explanation. First, independently 
\begin{figure*}
    \centering
    \includegraphics[width=\textwidth]{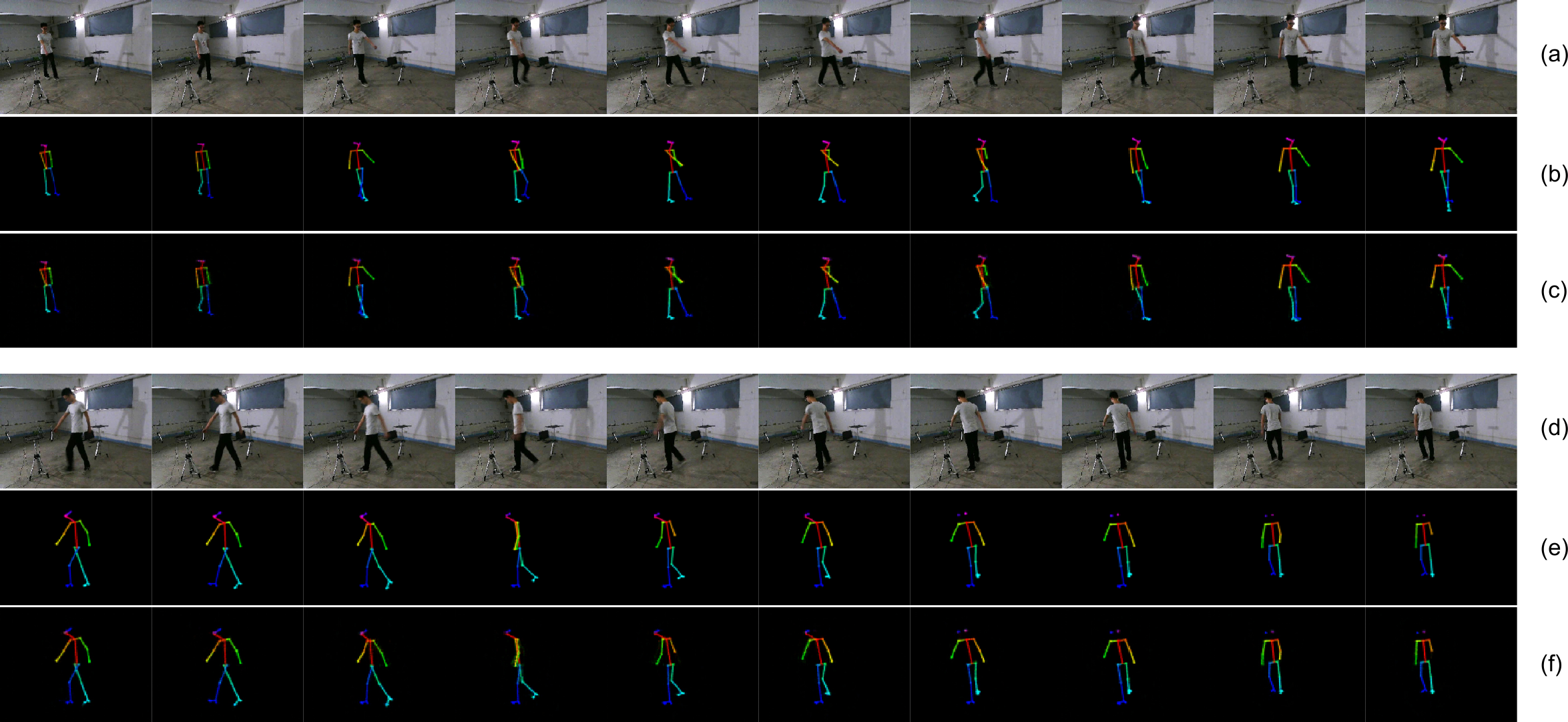}
    \caption{Test samples showing the human skeleton video synthesis for $h_P$ vector with the size of 100. In (a) and (d), the original RGB video frames are reported as visual reference. In (b) and (e), skeletons extracted from RGB frames representing the ground truth. Finally, in (c) and (f), the skeletons synthesized exploiting exclusively Wi-Fi signals.}
    \label{fig:results_skeleton}
\end{figure*}
\begin{figure*}
    \includegraphics[width=\textwidth]{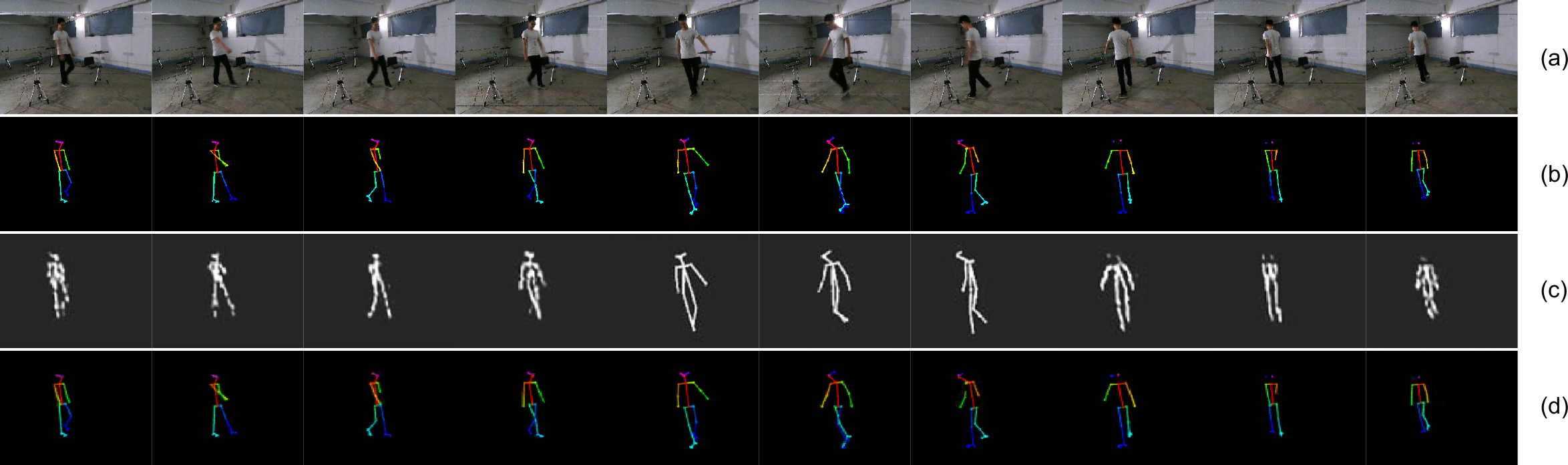}
    \caption{Qualitative comparison for the human skeleton synthesis. In (a) RGB frames as visual reference. In (b), the OpenPose generated ground truth. In (c) and (d), the human skeletons synthesized from Wi-Fi signals by Guo \textit{et al.} \cite{guo2020signal} and the proposed method, respectively.}
    \label{fig:comparison_skeleton}
\end{figure*}
\noindent
from vector $h_p$ size, enough information can be extracted from radio signals to correctly reconstruct skeleton videos. Second, the extracted radio features might be subject to noise when they have increased sizes, and it might affect the latent space representation, thus resulting in artifacts appearing inside synthesized videos, similarly to ghost silhouettes. This second aspect is caused by the OpenPose detailed skeleton, where bones connecting the various joints have different colors to indicate clearly, among other things, left from right body parts. As a matter of fact, this outcome is also supported by the

\begin{tablehere}
	\tbl{Latent signal-based feature vector $h_P$ size performance evaluation for human skeleton video synthesis. \label{tab:skeleton}}
	{\begin{tabular}{@{}cccc@{}}
		\toprule
		$h_P$ size & $MSE\downarrow$ & $SSIM\uparrow$ & $FSIM\uparrow$ \\ \colrule
		\textbf{100} & \textbf{0.001} & \textbf{0.954} & \textbf{0.991} \\
		200 & 0.003 & 0.770 & 0.964 \\
		300 & 0.005 & 0.776 & 0.949  \\
		400 & 0.001 & 0.944 & 0.989 \\
		\botrule
	\end{tabular}}
\end{tablehere}
higher number of epochs required for the student to reach good results on the skeleton video synthesis task. Regardless, when using a smaller $h_p$ size, the student can avoid extracting noise, therefore synthesizing correct skeleton videos starting from Wi-Fi signals. The latter can be observed in Fig.~\ref{fig:results_skeleton}, where the student model correctly reconstructs skeletons by considering a proper OpenPose color association.

In literature, due to the recent development of this field, Guo \textit{et al.}\cite{guo2020signal} are the sole authors currently performing experiments on the same, and only public collection available used to assess the proposed method on the skeleton image synthesis from Wi-Fi signals. Therefore, quantitative and qualitative comparisons with their work are reported to complete the human skeleton synthesis evaluation. Regarding the former, the evaluation was performed by computing the same custom metric devised by Guo \textit{et al.}, i.e., percentage of correct skeleton (PCS), to have a fair comparison. The obtained results are reported in Table~\ref{tab:pcs}. In detail, the PCS metric, which is inspired by the percentage of correct keypoint (PCK),  indicates the percentage of Euclidean distances between synthesized frames and their ground truths that lie within a variable threshold $\xi$. As shown, contrary to Guo \textit{et al.}, the student model achieves remarkable performances even for minimal threshold values, indicating the synthesis of exhaustive and high-quality skeleton frames. A result that can be likely associated with the architectural design that forces the decoder component to recreate accurate skeletons from Wi-Fi signals through cross-modality supervision. Regarding the qualitative comparison, synthesized skeletons are depicted in Fig. \ref{fig:comparison_skeleton}. As can be observed, even though both methods exploit OpenPose skeleton as ground truth, the presented approach synthesizes more accurate and less noisy skeletons, corroborating the results reported in Table 3. In fact, with respect

\begin{tablehere}
	\tbl{PCS metric comparison for different $\xi$ values, i.e., ground truth distance. Higher percentages correspond to a better synthesis quality.\label{tab:pcs}}
	{\begin{tabular}{@{}lcc@{}}
		\toprule
		Threshold & $Student$ & $Guo\;et al.$ \\  
		\colrule
		%$\xi = 0$ & 95.5\% & - \\
		$\xi = 1$ & 95.5\% & - \\
		$\xi = 3$ & 96.9\% & - \\
		$\xi = 5$ & 98.3\% & -  \\
		$\xi = 25$ & 100.0\% & 2.5\%  \\
		$\xi = 30$ & 100.0\% & 26.2\% \\
		$\xi = 40$ & 100.0\% & 75.6\% \\
		$\xi = 50$ & 100.0\% & 90.0\% \\
		\botrule
	\end{tabular}}
\end{tablehere}
to Guo \textit{et al.}\cite{guo2020signal}, the proposed model generates more consistent OpenPose skeletons that also take into account colors instead of binary maps, allowing to more easily identify the various limbs in the reconstructed image. Moreover, by synthesizing these detailed skeletons, the presented framework can also capture other details such as joints related to feet in the image. Such a result can be related to the extracted radio-based features that are mapped back to the visual domain by enforcing a similarity between the $Z$ and $V$ representations, fully highlighting and confirming the proposed cross-modality supervision and underlying architecture effectiveness. 

\section{Conclusion}
This paper presents a novel generative Wi-Fi sensing framework capable of synthesizing human silhouette and skeleton videos by exploiting exclusively wireless signals, eliminating privacy concerns in people monitoring and surveillance applications. The latter was achieved by designing a cross-modality learning strategy via a two-branch network that simulates a teacher-student model. Through this configuration, the architecture can focus on human body dynamics and build a mapping between different frequency spectra, i.e., visible and radio, by being trained on synchronized video-radio signal sample pairs. Most notably, the proposed two-branch network only requires visual data inputs at training time; then, by detaching the teacher model, the student can synthesize videos starting from wireless signals inputs. Since these signals are the only source of information for frame synthesis during the model evaluation, several ablation studies were performed on the low-dimensional radio features representation transferred into the visual domain to assess both silhouette and skeleton video synthesis. The results, obtained on a public dataset, indicate that the extracted radio features can influence the domain-to-domain mapping. Moreover, qualitative comparisons with other literature works highlight the effectiveness of the devised cross-modality learning approach since it enables the student network to synthesize more accurate and less noisy silhouette and skeleton videos.

As future work, a more challenging dataset will be collected to account for more elaborate human poses and more complex environments, where there is an increased signal interference and a higher number of people simultaneously present in the scene.
The former will be enable to evaluate the robustness of the proposed method in real-case scenarios where radio signal absorption, deformation, and superposition are common occurrences. The latter would instead open up additional Wi-Fi sensing applications of particular interest, where multiple people could be distinguished without video devices; enabling the recognition, for instance, of group actions from Wi-Fi signals. Moreover, further investigations will be performed on different signal properties in the time domain, e.g., impulse response or time of arrival, to predict human limb coordinates other than synthesizing visual representation. Finally, further inquiries on the skeleton synthesis will also be performed to address the pose estimation task from wireless signals. Specifically, the proposed framework will be adapted to implement a skeleton-based recognition component starting from the generated videos. This change will enable a joint investigation, on the one hand, of transverse approaches such as multi-task learning that can further refine the generated skeleton videos through extra information, on the other hand, of person verification capabilities from the generated video sequences that could be employed as an additional authentication tool in security scenarios.

\nonumsection{Acknowledgments}
\noindent This work was supported in part by the MIUR under grant “Departments of Excellence 2018–2022” of the Department of Computer Science of Sapienza University.

\bibliographystyle{ws-ijns}
\bibliography{sample}

\end{multicols}
\end{document}